\documentclass{ifacconf}

\usepackage{graphicx}      
\usepackage{natbib}        
\usepackage{amsmath}
\usepackage{amssymb}
\usepackage{tikz}
\usetikzlibrary{shapes,arrows}

\newtheorem{example}{Example}
\newtheorem{problem}{Problem}
\DeclareMathOperator*{\argmax}{arg\,max}
\DeclareMathOperator*{\argmin}{arg\,min}
\begin{document}
	\tikzset{%
		block/.style    = {draw, thick, rounded corners, rectangle, minimum height = 3em, 
			minimum width = 3em, text centered},
		sum/.style      = {draw, circle, node distance = 3cm}, 
	}

\newcommand{\suba}{\tiny$-$}
\newcommand{\suma}{\tiny$+$}
\newcommand{\abs}{$|\bullet|$}
\begin{frontmatter}

\title{Learning to falsify automated driving vehicles with prior knowledge} 


\author[First,Second]{Andrea Favrin} 
\author[First]{Vladislav Nenchev} 
\author[Second]{Angelo Cenedese}

\address[First]{BMW Group, 85716 Unterschleissheim, Germany (e-mail: \{andrea.favrin,vladislav.nenchev\}@bmw.de)}
\address[Second]{Department of Information Engineering, University of Padova, Italy (e-mail: angelo.cenedese@unipd.it)}

\begin{abstract}             
While automated driving technology has achieved a tremendous progress, the scalable and rigorous testing and verification of safe automated and autonomous driving vehicles remain challenging. This paper proposes a learning-based falsification framework for testing the implementation of an automated or self-driving function in simulation. We assume that the function specification is associated with a violation metric on possible scenarios. Prior knowledge is incorporated to limit the scenario parameter variance and in a model-based falsifier to guide and improve the learning process. For an exemplary adaptive cruise controller, the presented framework yields non-trivial falsifying scenarios with higher reward, compared to scenarios obtained by purely learning-based or purely model-based falsification approaches.
\end{abstract}


\end{frontmatter}

\section{Introduction}
\label{sec:introduction}

A challenge for developing automated and autonomous driving functions is the necessary rigorous testing for achieving safety and law compliance (\cite{Aptiv2019}), as well as a sufficient customer confidence level. Classical test approaches do not scale well for self-driving vehicles, as they require vast amounts of real-world driving to cover possible traffic situations. A promising approach for verification and validation of autonomous vehicles is testing in a virtual environment (\cite{Kim2017}). In addition to a sufficiently realistic simulation environment, a multitude of requirements have to be fulfilled to obtain a meaningful evaluation of a driving function from simulation. First, the allowed behavior of the autonomous vehicle must be defined in the form of an automatically verifiable specification, e.g., as a linear or metric temporal logic (LTL or MTL) formula (\cite{Alur1992}) or an unambiguously ordered set of rules (\cite{Censi2019}) akin to a violation metric on possible scenarios. Second, the executed test scenarios must cover typical cases, as well as rare but realistic driving situations. Data-driven approaches provide a remedy to a certain extent (\cite{Eggers2018}), but analyzing large amounts of collected real data cannot guarantee that all relevant scenes are included. Third, a systematic perturbation of sensor data has to be performed, if machine learning components are used, unless safety properties are embedded by design (\cite{Nenchev2019}). Forth, the System Under Test (SUT) should be the actual implementation of the function, consisting of as many of the involved software and hardware components as possible. Many approaches exist that fulfill subsets of these requirements, typically focusing on parts of the overall system, comprising perception, planning and control components. On one hand, simulation-based verification approaches have been proposed that perform sampling-based testing for estimating accident probability under standard traffic behavior (\cite{OKelly2018}), or falsifying a formal specification for the closed-loop system in the presence of environment uncertainty for perception components (\cite{Tommaso2019}). In these approaches, the simulator and the SUT are treated as a black-box, such that generating meaningful falsifying scenarios might require millions of computationally expensive simulations. On the other hand, model-based approaches have been successfully used for falsification, e.g., finite automata-based abstractions for direct model checking (\cite{Volker2019}), as well as simplified analytical models in local optimization (\cite{Althoff2018}), sampling-based search (\cite{Koschi2019}) or global optimization (\cite{Tuncali2017}) for falsifying, among others, Adaptive Cruise Control (ACC) controllers. While these approaches can be computationally advantageous or even offer completeness guarantees with respect to the employed model, generalizing the results to the actual implementation of driving function is not automatically given.

In this paper, we propose a learning-based framework for falsifying the implementation of an automated or self-driving function in simulation. Instead of only focusing on safety properties, we aim at falsifying its complete specification, given with an associated violation metric for possible scenarios. Building upon the concept of adversarial agents, falsification is addressed by an Adversarial Reinforcement Learning Agent (ARLA) that chooses scenario parameters to maximize the specification violation metric. Even though Reinforcement Learning (RL) approaches may lead to a good performance when a traditional model of the system is hard to obtain (\cite{Luong2019}), they come at the cost of high variance in actions leading to slow learning. Despite the tremendous increase of available computational power, simulations might be expensive to perform, in particular, when realistic sensor inputs, e.g. camera images, lidar point clouds etc., have to be generated. As a first measure, we utilize a model of the dynamical behavior of the controlled vehicle in its environment to limit the action variance to reasonable scenarios. Even though this is shown to improve the learning progress significantly in the provided case study of falsifying an adaptive cruise controller, the scenario parameter state space size becomes the main limiting factor for falsifying more complex driving functions. Therefore, as a second measure, ARLA is augmented by a model-based falsifier. Using a controlled vehicle model, an estimate of the reward for a scenario can be obtained. Thus, ARLA's reward is modified such that it is large only if the actual reward is greater than the estimated reward. The overall scenario parameter is obtained by combining the outputs of the model-based falsifier and ARLA. The effectiveness of the presented learning-based framework is demonstrated for falsifying an exemplary adaptive cruise controller. A comparison to a pure model-based and a pure learning-based approach shows that the method yields non-trivial falsifying scenarios with higher reward.

The remainder of the paper is organized as follows: In Sec.\,\ref{sec:problem_formulation}, the problem formulation and the specification of ACC as a running example are provided. Then, in Sec.\,\ref{sec:solution}, we present the baseline ARLA, which is then extended to the learning-based falsification framework with prior knowledge. In Sec.\,\ref{sec:results}, we provide empirical results, followed by a discussion and conclusions in Sec.\,\ref{sec:conclusions}.

\section{Problem formulation}
\label{sec:problem_formulation}

Consider a host vehicle with state $x_h\in X_h\subseteq \mathbb{R}^{n_h}$ moving in an environment with state $x_e\in X_e\subseteq \mathbb{R}^{n_e}$. The vehicle is equipped with sensors that provide measurement data of the environment $y_h \in Y_h$, denoted by a mapping $O:X_e\to Y_h$. Assuming that the vehicle is equipped with an automated or autonomous driving function, its dynamics over time $t\in[0,T]$ are described by the system
\begin{equation}\label{eq:host}
 x_{h,t+1}=f_h(x_{h,t}, y_{h,t}).
\end{equation}
We adopt an adversarial perspective of the environment, i.e., the environment has complete knowledge of the host vehicle state, and evolves according to 
\begin{equation}\label{eq:env}
 x_{e,t+1}=f_e(x_{e,t}, x_{h,t}, u_{e,t}),
\end{equation}
where $u_e\in U_e$ denotes a finite set of parameters. Both mappings $f_h$ and $f_e$ are assumed to be deterministic. Let $X=X_h\times X_e$ denote the overall state space of the system, comprising the host vehicle and its environment, and $x_{[0,T]}$ a finite trace of the system over time $t\in[0,T]$. Since our goal is to falsify the automated or autonomous driving function, the (adversarial) parameter space $P$ contains all environment parameters as well as the complete initial state $x_0\in X$, i.e., $P=X\times U_e$.

Consider a specification $\varphi$ for the controlled host vehicle given as a finite set of properties associated with a violation metric over possible scenarios. Let a finite time trace $x_{[0,T]}$ of the system that satisfies the specification $\varphi$ be formally denoted by $x_{[0,T]}\vDash\varphi$. 

\begin{example}[Keep distance to a front vehicle]
Consider an ACC function. Introducing the time headway $t_{h}$, defined as the distance $h$ over the velocity $v$, i.e., $t_h=h/v$, the ACC requirements are:
\begin{itemize}
 \item possible modes: \textit{set speed mode} and \textit{time gap mode};
 \item in \textit{set speed mode}, a driver desired speed $v_d\in[v_{d_{min}},v_{d_{max}}]$ eventually needs to be maintained;
 \item in \textit{time gap mode}, a desired time headway $t_{h_d}$ to the front vehicle eventually needs to be maintained, and the time headway $t_h$ needs to satisfy $t_h\in[t_{h_{min}},t_{h_{max}}]$ at all times;
 \item The system is in \textit{set speed mode}, if $v_d\leq h/t_{d}$, otherwise it is in \textit{time gap mode};
 \item the acceleration $a_h\in[a_{h_{min}},a_{h_{max}}]$ at all times.
\end{itemize}
\end{example}

The problem of falsifying the controlled host vehicle can be stated as follows.
\begin{problem}
Given an automatically verifiable specification $\varphi$ for the controlled host vehicle \eqref{eq:host} in its environment \eqref{eq:env}, find a parameter $p\in P$ yielding a finite trace $x_{[0,T]}\not\vDash\varphi$.
\end{problem}

From a formal methods perspective, the problem above represents a counterexample search. Simplified versions of this problem might be amenable for model checking approaches, if analytical models for \eqref{eq:host} and \eqref{eq:env} exist and the dimensionality of the parameter space $P$ can be reduced. However, we are interested in falsifying the actual implementation of the automated driving function, possibly even executed on the embedded electronic control unit, to also take into account real-time behavior. For that, we can only resort to simulation-based verification. Further, we want a scalable approach that can easily adapt during any development phase of the function and not only at the end of a release cycle. Our main goal is to obtain qualitatively interesting falsification scenarios that go beyond the scope of typical human-experience-derived test cases, and not complete verification of the function. To fulfill all of these requirements, the problem is addressed by an adversarial reinforcement learning agent.

\section{Learning-based solution}
\label{sec:solution}

The proposed solution is based on learning to falsify the automated or self-driving function by generating driving scenarios, in which the controlled host vehicle violates its specification. The agent repeatedly runs simulations of the controlled host vehicle \eqref{eq:host} and its environment \eqref{eq:env} to observe the state of the system, receive a reward, and improve the scenario parameters for the next run. To improve the learning behavior, we present how prior knowledge models can be incorporated to limit the scenario parameter variance, or to augment the agent by a model-based falsifier. We start with the (pure learning-based) baseline agent.

\subsection{Baseline adversarial agent}\label{sec:baseline}

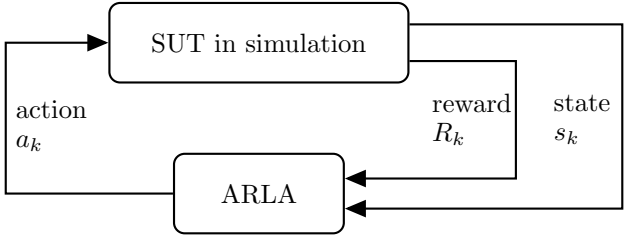
\begin{figure}
	\centering
	
	\begin{tikzpicture}[auto, thick, node distance=1cm, >=triangle 45]
	\node (environment) [block,  text width=3.7cm] {SUT in simulation};
	\node (agent) [block, below of=environment, yshift=-1cm, text width=2cm]
	{ARLA};     
    \draw[->] (environment.7) -- ++(2.8,0) node[xshift=-0.4cm,yshift=-0.8cm, text width=1cm, align=left, below]	{state $s_k$} |-(agent.-10);
	\draw[->] (environment.-7) -- ++(1.4,0)  node[xshift=-0.6cm,yshift=-0.3cm, text width=1cm, align=left,below]
	{reward $R_k$}|- (agent.10);
	\draw[->] (agent.west) -- ++(-2.2,0) node[yshift=0.9cm, text width=1cm,align=left,above, right]
	{action $a_k$}|-(environment.west);
	\end{tikzpicture}
	\caption{Baseline falsification with adversarial agent.} 
	\label{fig:simplerlagent}
\end{figure}

To obtain the simulated system trace $x_{[0,T]}$, the agent takes an initial action $a_{0}\in A$, which is mapped to a scenario parametrization $p$ by $f_{act}: A\to P$, i.e., $p=f_{act}(a_0)$. Given a time-invariant evaluation mapping over a trace of the system $f_{eval}: x_{[0,T]}\to S$, we compute the state $s_k\in S$ that comprises relevant variables for checking if the trace satisfies the specification. At every simulation run, the agent receives a reward captured by $R: S \times A\to \mathbb{R}_+$. Even though a trace $x_{[0,T]}$ satisfying the specification $\varphi$ is a binary property, under mild assumptions, we assume the existence of a function $\rho:x_{[0,T]}\to \mathbb{R}$ that provides a violation metric of the specification on possible scenarios for the system. Assuming an LTL or MTL specification, a real-valued function $\rho^\varphi(s,a)$ can be readily associated with the quantitative semantics of the formula as shown in \cite{Fainekos2009}. When the simulated system trace $x_{[0,T]}$ violates the specification, the violation metric is $\rho^\varphi(s,a)<0$, and $\rho^\varphi(s,a)\geq 0$ otherwise. Thus, to facilitate learning, the reward function is chosen as
\begin{equation}\label{eq:reward}
R(s,a)= exp(-\rho^{\varphi}(s,a))),
\end{equation}
such that the agent receives a high reward for a trace that violates the specification. Using the collected rewards, the agent performs re-training, and produces a new action that triggers a new simulation, as shown in the scheme (Figure~\ref{fig:simplerlagent}). If additional constraints for generating reasonable scenario parameters of the form
\begin{align}\label{eq:scenarioconstraints}
 \begin{aligned}
  g_{sc}(s,a)=0,\\
  h_{sc}(s,a)\geq 0,
 \end{aligned}
\end{align}
exist, the reward function is additively augmented by corresponding terms.
\begin{example}
For Ex.\,1, the action $a_k$ consists of the initial host vehicle velocity $v_{0}$, initial distance to the front vehicle $d_{0}$ and a traffic vehicle velocity profile, encoded by a finite time series of piecewise-constant acceleration segments $a_{f_n},t_n \in {t_0,t_1,\ldots,t_N}$. The state $s_k$ contains the minimal and maximal host acceleration $a_{\text{min}}$ and $a_{\text{max}}$, the minimum time headway $t_{h,min}$ and the minimum and maximum velocities $v_{\text{min}}$ and $v_{\text{min}}$ over the trace $x_{[0,T]}$. 

Consider the state sets for the ACC specification (Ex.\,1), denoting the vehicle in set speed mode $M_1=\{(x_h,u_h): v_d\leq h/t_{h_d}\}$ with its safe set $S_1=X_h\times U_h$ and target set $T_1=\{(x_h,u_h):v\leq v_d\}$, and the set denoting time gap mode $M_2=\{(x_h,u_h): v_d> h/t_{h_d}\}$, its safe set $S_2=\{(x_h,u_h):v\leq h/t_{h_{min}}\}$ and target set $T_2=\{(x_h,u_h):v\leq h/t_{h_d}\}$, as well as the overall safe set $S_U=\{(x_h,u_h): a_h\in[a_{h_{min}},a_{h_{max}}]\}$. Then, the LTL specification for ACC (\cite{Nilsson2016}) is given by
\begin{align}
\begin{aligned}
 \varphi=\mathbf{G} (S_U \land (\land_{i=1}^2((M_i{\implies} S_i)\land (\mathbf{G}M_i {\implies} \mathbf{GF} T_i)))),
\end{aligned}
\end{align}
where $\mathbf{G}$ and $\mathbf{F}$ denote the temporal operators ``globally'' and ``eventually'', respectively. The LTL formula is associated with quantitative semantics (\cite{Fainekos2009}) to obtain \eqref{eq:reward}. In addition, scenario constraints \eqref{eq:scenarioconstraints} exist, requiring that the sum of all time segments of the acceleration of the front vehicle must be equal to the scenario duration $t_{dur}$, i.e., $\sum_n t_n -t_{dur}=0$.
\end{example}

While RL may lead to a good performance when a traditional model of the system is hard to obtain, complex tasks might take millions of iterations until meaningful actions are generated. High variance of the produced action remains a significant hurdle for RL applications and turn learning parameters tuning into a tedious task. To mitigate these effects, in the following, we incorporate prior knowledge of the automated driving vehicle in its environment to focus learning.

\subsection{Limiting action variation with prior knowledge}\label{sec:limit}

To improve learning performance, the action variation is restricted to reasonable scenarios by prior knowledge. Since our goal is to obtain qualitatively interesting falsification scenarios, any scenario that violates the specification in a foreseeable manner could be excluded. A trivial specification violation emerges, e.g., when the scenario starts in an initial state, where a collision of the controlled host vehicle is unavoidable, assuming typical host and environment dynamics. Another example for a trivial specification violation is, when the scenario duration is too short for the host vehicle to reach the driver desired velocity $v_{\text{d}}$. Consider an analytical model of the host and the environment's dynamics
\begin{equation}\label{eq:estmodel}
\hat{x}_{t+1}=\hat{f}(\hat{x}_t,a_k,\hat{u}_t),
\end{equation}
where $\hat{u}_t$ denotes the control input of the host vehicle. Based on \eqref{eq:estmodel}, a trivial specification violation can be checked for a given host control input $\hat{u}_t$ and a given action $a_k$ (mapping to a corresponding scenario parametrization $p_k$). Then, assuming that the initial state of the system $\hat{x}_0$ is contained in $p_k$, the host vehicle starting in a specification non-violating (safe) state can be captured by the inequality 
\begin{equation}\label{eq:ineqavoid}
g(a_k,\hat{u}_t)\leq 0.
\end{equation}
To reduce the action variance of ARLA, if the output of the agent $a_{nn}$ violates \eqref{eq:ineqavoid}, the action is projected to the boundary of the inequality. This corresponds to solving the optimization problem
\begin{align}\label{eq:projection}
\begin{aligned}
 a_k=&\argmin_{a\in A} \|a_{nn}-a\|\\
 \text{s.t. } & \hat{x}_{t+1}=\hat{f}(\hat{x}_t,a,\hat{u}_t), t\in[0,T],\\
 &g(\hat{x}_{[0,T]},a)= 0,
 \end{aligned}
\end{align}
where an appropriate control input $\hat{u}_t$ has to be chosen for the particular scenario class.
\begin{example}
For the ACC example, consider only longitudinal motion and, thus, assume a vehicle model with state $x_{(\cdot),k}=[s_{(\cdot),k},v_{(\cdot),k}]^T$, sampling time $t_s$ and
\begin{align}\label{eq:vehmodel}
\begin{aligned}
s_{(\cdot),k+1}&=s_{(\cdot),k}+ t_s v_{(\cdot),k} + 0.5 t_s^2 u_{(\cdot),k}\\
v_{(\cdot),k+1}&=v_{(\cdot),k}+t_s u_{(\cdot),k},
\end{aligned}
\end{align}
where $(\cdot) =\{h,f\}$ denotes the host and the front vehicle, respectively. In this simple case, the overall model \eqref{eq:estmodel} is given by one instance of \eqref{eq:vehmodel} for the host and one for the front vehicle. Given the ACC specification (Ex.\,1), the function should ensure collision-free motion of the vehicle at all times. In this scenario class, the worst case occurs when the front vehicle performs an emergency braking with acceleration $a_{f,min}$, i.e., $u_{f,k}=a_{f,min}$. Thus, for the host vehicle, we can also assume maximal braking with acceleration $a_{h,min}$ as an input, i.e., $\hat{u}_t=a_{h,min}$. Then, the braking distances $s_{stop, h}$ and $s_{stop, f}$ can be computed by substituting the corresponding values in \eqref{eq:vehmodel}. A collision occurs, if the stopping distance of the host vehicle without the desired distance $d_{fh}$ between the two vehicles is larger than the stopping distance of the front vehicle, i.e.,
\begin{equation*}
s_{stop, h}-d_{fh} \geq s_{stop, f}.
\end{equation*}
Substituting the corresponding equations \eqref{eq:vehmodel} for the host and the front vehicle, for \eqref{eq:ineqavoid} we obtain
\[
s_{h,0} - \frac{v_{h, 0}^2}{2 a_{h, min}}-d_{fh} \geq s_{f,0} - \frac{v_{f,0}^2}{2 a_{f,min}},
\]
which can be used to ensure that the vehicle starts a simulation run in a specification non-violating (safe) state. Similarly, to check for reaching the driver desired velocity, the maximum comfortable acceleration can be assumed for deriving a corresponding inequality.
\end{example}

\subsection{Combining model-based falsification and learning}\label{sec:combined}

Despite the substantial learning improvement by excluding trivially violating parametrizations as described in Sec.\,\ref{sec:limit}, it may still take a large number of executions until interesting scenarios are generated by ARLA. Thus, we additionally introduce a model-based falsifier. In addition to the host vehicle's model \eqref{eq:estmodel}, let a host controller model be given by
\begin{equation}\label{eq:estcontrol}
\hat{u}_t=h(\hat{x}_t,a_k).
\end{equation}
In the best case, \eqref{eq:estcontrol} should be the actual algorithm used in the SUT, but it can also be a simplified controller that can be executed and evaluated quickly to avoid computationally expensive environment simulations. 

\begin{example} 
For ACC, let the overall model \eqref{eq:estmodel} contain vehicle models given by \eqref{eq:vehmodel} for each of the host and front vehicles, with corresponding variable subscripts $h$ and $f$, respectively. As a host controller model \eqref{eq:estcontrol}, we assume the Intelligent Driver Model (\cite{Treiber2000}):
\begin{align}\label{eq:acccontroller}
\begin{aligned}
&\hat{u}_t(\hat{x}_t,a_k)=a_{h,max}(1-(\frac{v_{h,t}}{v_d})^\alpha-(\frac{d(s_{h,0},v_{h,t},v_{f,t})}{d_{fh}})^2)\\
&d(s_{h,0},v_{h,t},v_{f,t})=s_{h,0}+v_{h,t} t_{h_d}+\frac{v_{h,t} (v_{h,t}-v_{f,t})}{2\sqrt{a_{h,max} a_{h,com}}},
\end{aligned}
\end{align}
where $d_{fh}$ is the desired distance between the two vehicles, $a_{h,max}$ and $a_{h,com}$ are the absolute values of the maximally allowed and the comfortable accelerations, respectively, and $\alpha$ is a tuning parameter. 
\end{example}

Based on the dynamical models \eqref{eq:estmodel}, the host controller \eqref{eq:estcontrol}, and the action variation constraints \eqref{eq:ineqavoid}, the model-based falsifier is formulated as the optimization problem:
\begin{align}\label{eq:modelfalsifier}
\begin{aligned}
\hat{a}_k=\argmax_{a\in A, s\in S} & \eqref{eq:reward},\\
\text{s.t. }& \eqref{eq:scenarioconstraints},\eqref{eq:estmodel}, \eqref{eq:ineqavoid}, \eqref{eq:estcontrol}, 
\end{aligned}
\end{align}
for determining an estimate of the optimal falsifying action $\hat{a}_k$ that corresponds to an estimate of the maximal reward $\hat{R}_k$ in the $k$-th iteration. As initial optimization states $a_{init},s_{init}$, we take the action and the state of the previous run $a_{k-1}, s_{k-1}$, respectively. Since the action $a$ has only an influence on the initial condition of the optimization problem, we can compute the gradient $d R(a,s)/d a$ by substituting \eqref{eq:estcontrol} into \eqref{eq:estmodel}, and then, substitute \eqref{eq:estmodel} into the reward function \eqref{eq:reward}, and compute the solution of the optimization problem $\hat{a}_k$, e.g., by a gradient descent approach. Since the optimization problem is typically (heavily) non-convex, we can resort to using global optimization algorithms. Note that the approaches proposed in, e.g., \cite{Tuncali2017} or \cite{Koschi2019}, are possible realizations of the model-based falsifier \eqref{eq:modelfalsifier}.


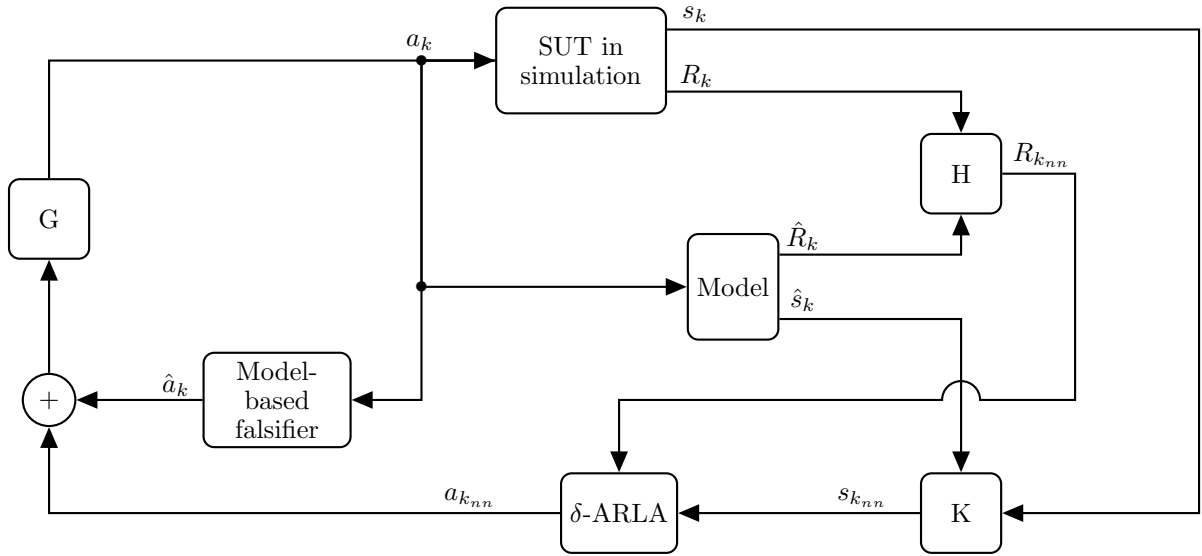
\begin{figure*}[t]
	\centering
	\begin{tikzpicture}[thick, node distance=3cm, >=triangle 45]
	\node (environment) [block, minimum height = 1.4cm, text width = 2cm] {SUT in simulation};
	\node (optimizer) [block, below of=environment, xshift=-4cm, yshift=-1.5cm, text width=1.7cm] {Model-based falsifier};
	\node (projection) [block, left of=optimizer, yshift= 2.4cm] {G};
	\node (model) [block, below of=environment, xshift= 2cm, minimum height = 1.4cm] {Model};
	\node (reward) [block, right of=environment, xshift=2cm, yshift=-1.5cm] {H};
	\node (difference) [block, below of=reward,  yshift=-1.5cm] {K};
	\node (agent) [block, left of=difference, xshift=-1.5cm] {$\delta$-ARLA};
	
	\node (sum) [sum, left of=optimizer] { + };
	\draw[->] (environment.-20) -| node[xshift=-3.2cm, yshift  =0.21cm, text width=1cm] {$R_k$} (reward);
	\draw[->] (environment.20) -- ++(7,0) node[xshift=-6.3cm, yshift  =0.2cm, text width=1cm] {$s_k$} |- (difference.east);
	\draw[->] (model.35) -| node[xshift=-1.8cm, yshift  =0.25cm, text width=1cm] {$\hat{R}_{k}$} (reward);
	\draw[->] (model.-35) -| node[xshift=-1.75cm, yshift  =0.25cm, text width=1cm] {$\hat{s}_{k}$} (difference);
	\draw[->] (reward) -- ++ (1.5, 0) node[xshift=-0.32cm, yshift  =0.25cm, text width=1cm] {$R_{k_{nn}}$} -- ++(0, -3) -- ++ (-1.25, 0) arc(0:180:0.25cm) -| (agent);
	\draw[->] (difference) -- node[xshift=1cm, yshift  =0.2cm, text width=1cm] {$s_{k_{nn}}$} (agent);
	\draw[->] (agent) -| node[xshift=5.7cm, yshift  =0.2cm, text width=1cm] {$a_{k_{nn}}$} (sum);
	\draw[->] (environment) -- ++(-2.1, 0) node[xshift=0.3cm, yshift  =0.24cm, text width=1cm] {$a_{k}$} |- (model);
	\draw[->] (environment) -- ++(-2.1, 0) |- (optimizer);
	\draw[->] (sum) -- (projection);
	\draw[->] (optimizer) -- node[xshift=0.8cm, yshift  =0.2cm, text width=1cm] {$\hat{a}_{k}$} (sum);
	\draw[->] (projection) |- (environment);
	\draw node at (-2.1,-0.01){\textbullet};
	\draw node at (-2.1,-3.01){\textbullet};
	\end{tikzpicture}
	\caption{Combined scheme of model-based falsifier and $\delta$-ARLA.}
	\label{fig:combinedscheme}
\end{figure*}

The model-based falsifier \eqref{eq:modelfalsifier} generates a counter-example that represents the best guess under the assumed models \eqref{eq:estmodel} and \eqref{eq:estcontrol}. Thus, the remaining goal for the adversarial agent is to learn the behavioral difference between the model and the SUT to generate better falsifying actions. Figure~\ref{fig:combinedscheme} shows the scheme for combining the model-based falsifier and the corresponding adversarial agent, which will be denoted by $\delta$-ARLA. Based on the models \eqref{eq:estmodel} and \eqref{eq:estcontrol}, an estimate of the state $\hat{s}_k$ and the reward $\hat{R}_k$ can be computed. Then, in parallel to the model-based falsifier, the state of $\delta$-ARLA is given by
\begin{align}  
K: s_{k_{nn}}=s_k-\hat{s}_k.
\end{align}
Since the goal of $\delta$-ARLA is to learn the difference between the model and the actual system, $\delta$-ARLA's reward is chosen such that the $R_{k_{nn}}$ is high only when the actual reward $R(s_k,a_k)$ is high and greater than the estimated reward $\hat{R}_k$. The reward can be written as  
\begin{align}  \label{eq:deltareward}
H: R_{k_{nn}}=R(s_k,a_k)+max(0,R(s_k,a_k)-\hat{R}_k).
\end{align}
Finally, we compute the combined action as $\tilde{a}_{k}=\hat{a}_{k-1}+(\hat{a}_{k-1}-a_{k_nn})+(\hat{a}_{k-1}-\hat{a}_k)$. To focus learning, the combined action $\tilde{a}_{k}$ is limited as described in Section~\ref{sec:limit} to exclude trivially violating scenarios, i.e., $a_k=G(\tilde{a}_k)$.

\subsection{Learning architecture}

The proposed approach is amenable for any RL model, capable of handling continuous variables. In particular, we utilize a deep neural network as a prediction model for ARLA and $\delta$-ARLA due to its empirically proven beneficial modelling performance for complex systems and large amounts of data. We employ Deep Deterministic Policy Gradient (DDPG), because it has been shown to handle continuous and high-dimensional action spaces well and is particularly suitable for deterministic functions (\cite{Lillicrap2016}).

\section{Simulation results}
\label{sec:results}

We apply the proposed learning-based framework for falsifying an implementation of the ACC controller \eqref{eq:acccontroller} with respect to the formal specification derived in Example\,2. In addition to comparing the learning-based schemes presented in Sec.\,\ref{sec:solution}, we also assess their performance with respect to a purely model-based falsifier.

A discrete-time version of \eqref{eq:acccontroller} is considered as $u_{h,k}$ in the SUT. For a more realistic simulation of the vehicle behavior controlled by the SUT, we augment the host vehicle dynamics \eqref{eq:vehmodel} by a friction coefficient $\alpha=-0.01$, i.e., $v_{h,k+1}=(1-\alpha)v_{h,k}+t_s u_{h,k}$, where $t_s=0.1~s$. Note that as a prior knowledge model for computing the cost and state estimates $\hat{R}_k$ and $\hat{s}_k$, and in the model-based falsifier \eqref{eq:modelfalsifier}, we assume the simpler dynamics \eqref{eq:vehmodel}. 
The host vehicle is assumed to have the maximal allowed acceleration and deceleration according to the \cite{ISO15622} standard, i.e. $a_{h_{min}} = -3.5\,m/s^2$ and $a_{h_{max}} = 2\,m/s^2$. The front vehicle is assumed to have the typical maximum acceleration and deceleration, i.e. $a_{f_{max}} = -0.8\text{g}\,m/s^2$ and $a_{f_{min}} = 0.4\text{g}\,m/s^2$ with the gravitational constant $\text{g} = 9.82\,m/s^2$. The behavior of the front vehicle is given over $n=5$ constant acceleration segments $a_{f_n}$, as described in Example~2. The scenario duration is fixed to $t_{dur}=30 s$. The reward for a simulated trajectory is obtained based on the violation score of the LTL properties obtained by the py-metric-temporal-logic-package (\cite{Tommaso2019}). The optimization problem corresponding to the model-based falsifier \eqref{eq:modelfalsifier} for the combined scheme (Figure~\ref{fig:combinedscheme}) is solved by the BFGS local optimizer in SciPy. For the purely model-based falsifier, we employ the global optimization (SHGO) solver in SciPy. For ARLA, a deep neural network is used, where both the actor and the critic have two hidden layers with corresponding sizes for the defined action and state spaces. Simulating and evaluating a particular scenario with the chosen parameters takes $0.3~s$ on an Intel Core i7-8565U. Training the neural network with a batch size of 128, a discount factor 0.99 and learning rates of $0.0003$ for both the actor and the critic, takes $30\,s$ for an episode of the pure learning-based approach on an NVidia GeForce Quadro M2000M GPU. Episode convergence time decreases by a factor of $2$ for the combined approach.

\begin{figure}
	\includegraphics[width=0.5\textwidth]{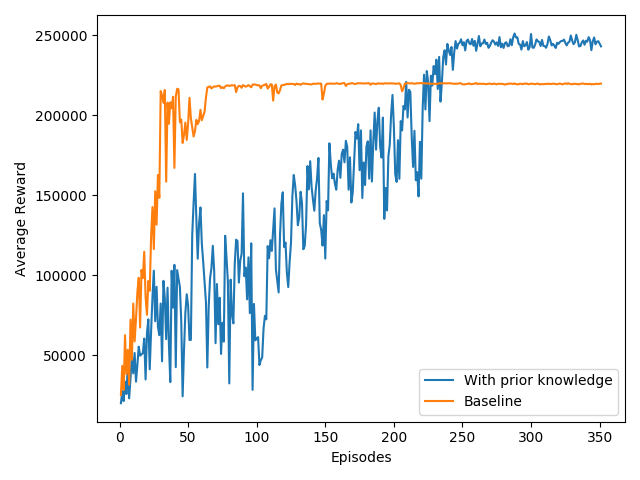}
	\caption{Learning progress curve: comparison between baseline and prior knowledge}
	\label{fig:comparison}
\end{figure}

\begin{figure}
	\includegraphics[width=0.5\textwidth]{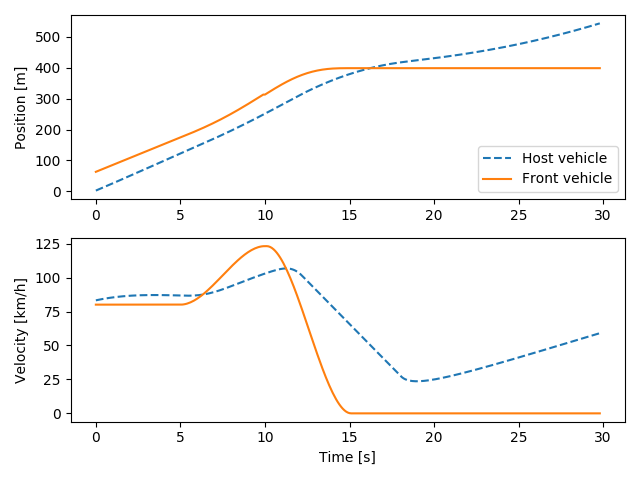}
	\caption{Simulation of the best falsifying scenario with learning-based falsification with prior knowledge.}
	\label{fig:trajectory}
\end{figure}

\begin{figure}
	\includegraphics[width=0.5\textwidth]{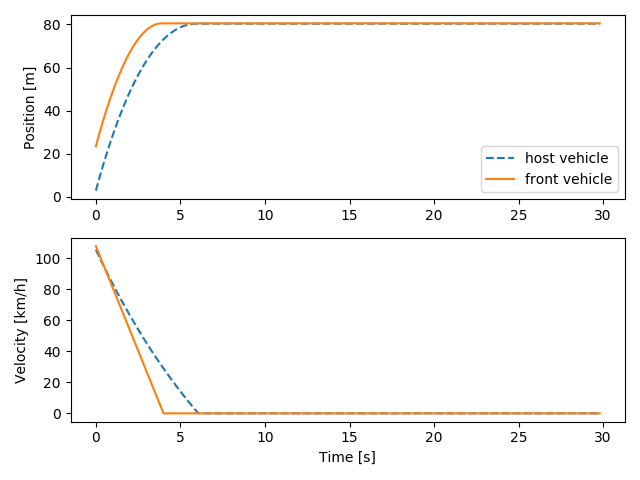}
	\caption{Simulation of the best falsifying scenario with purely learning- or purely model-based falsification.}
	\label{fig:trajectory1}
\end{figure}

Figure~\ref{fig:comparison} shows the learning progress of the purely learning-based baseline approach (Section~\ref{sec:baseline}) compared to learning with limiting action variation (Section~\ref{sec:limit}) over 350 episodes. Limiting the action with prior knowledge not only helps to find scenarios with a higher reward, but the learning progress is more gradual increasing over the epochs, compared to the baseline case. A similar learning and reward progress is also reached with the combined approach. In the particular case, the reward deviation \eqref{eq:deltareward} is small due to the small difference of the friction coefficient in the actual reward.
Figure~\ref{fig:trajectory} shows a simulation of the scenario resulting from the learning-based framework with prior knowledge. The scenario is a non-trivial falsification of the ACC specification, since a collision happens around $16\,s$, even though the host vehicle starts in a safe state, i.e., a state where a collision can be physically prevented by an appropriate control action. Interestingly, both the baseline learning-based solution and the purely model-based falsifier converge to an ``expected'' falsifying scenario, as shown in Figure~\ref{fig:trajectory1} -- a collision occurs, when the front vehicle immediately performs a full braking with maximal allowed deceleration, while the host vehicle starts in a safe state. Note that the corresponding reward is lower than the reward found by our learning-based approach, even though the same initial states were used for all schemes. This is a consequence of the non-convexity of the reward function, which makes finding the global maximum very hard. More importantly, it shows that combining prior knowledge with learning leads to an improved falsification outcome.

\section{Discussion and conclusions}
\label{sec:conclusions}

We presented a learning-based approach for the automatic generation of scenarios that falsify the implementation of an automated or self-driving function in simulation. Our approach relies on efficiently combining prior knowledge models with reinforcement learning. This allows  developers  to  detect  flaws in the system at any stage of the development progress. Our approach outperforms both pure learning-based and pure model-based methods by generating non-trivial scenarios that falsify the considered controller. The method can be readily extended for falsifying larger parts of (more complex) driving functions. While it requires running many simulations (possibly executed in parallel) initially, our experiments show that it provides a critical situation that would otherwise require several thousands of kilometers of real, or ``randomized'' scenario driving in a simulator. Unfortunately, none of the existing falsification approaches can guarantee completeness with respect the actual implementation of the system, i.e., even when no falsifying scenario is found, there is no guarantee that such a scenario does not exist. This could be addressed by quantifying the quality of the obtained learned model with respect to base distribution simulation models for standard behavior of other traffic participants, weather conditions etc., as a subject of future work.

\bibliography{rlverification}             
\end{document}